\documentclass{article}
\usepackage{spconf,amsmath,graphicx}

\usepackage{epsfig}
\usepackage{graphicx}
\usepackage{amsmath}
\usepackage{amssymb}
\usepackage[ruled,vlined,noresetcount]{algorithm2e}
\SetKwInOut{Parameter}{Parameters}
\SetKwData{Left}{left}
\SetKwData{This}{this}
\SetKwData{Up}{up}
\SetKwFunction{Union}{Union}
\SetKwFunction{FindCompress}{FindCompress}
\SetKwInOut{Input}{input}
\SetKwInOut{Output}{output}
\SetKwInput{KwInput}{Input}                % Set the Input
\SetKwInput{KwOutput}{Output}              % set the Output
\title{Collaborative Learning to Generate Audio-Video Jointly}
\vspace{-2mm}
\name{Vinod K Kurmi$^{\star}$ \hspace{-0.5em}\quad Vipul Bajaj$^{\star}$ \hspace{-0.5em}\quad Badri N Patro$^{\star}$ \hspace{-0.5em} \quad K S Venkatesh$^{\star}$ \hspace{-0.5em} \quad Vinay P Namboodiri$^{\dagger}$ \hspace{-0.5em} \quad Preethi Jyothi$^{\ddagger}$\vspace{-4mm}}
\vspace{-15em}
\vspace{-2mm}
			\address{\normalsize{$^{\star}$ Indian Institute of Technology Kanpur \quad
			    $^{\dagger}$University of Bath
			    \quad
			    $^{\ddagger}$Indian Institute of Technology Bombay}\vspace{-4mm}}

\begin{document}
%\ninept
%
\maketitle
\begin{abstract}
 There have been a number of techniques that have demonstrated the generation of multimedia data for one modality at a time using GANs, such as the ability to generate images, videos, and audio. However, so far, the task of multi-modal generation of data, specifically for audio and videos both, has not been sufficiently well-explored. Towards this, we propose a method that demonstrates that we are able to generate naturalistic samples of video and audio data by the joint correlated generation of audio and video modalities. The proposed method uses multiple discriminators to ensure that the audio, video, and the joint output are also indistinguishable from real-world samples. We present a dataset for this task and show that we are able to generate realistic samples. This method is validated using various standard metrics such as Inception Score, Frechet Inception Distance (FID) and through human evaluation.
\end{abstract}
\vspace{-1em}

%\vspace{-1.01em}
\section{Introduction}
\vspace{-0.701em}
% Data is the foremost requirement for any machine learning task which is further escalated in case of deep learning. Data, if used in the right direction can be extremely powerful and plays a key role of any decision. American statistician, W. Edwards Deming once famously said,
% \begin{center}
% \textit{In God we trust. Everyone else, bring data.}\\    
% \end{center}

% But data can be of varied types, and real-life data is inherently multimodal. 

\begin{figure}[!t]
	\centering
	\includegraphics[width=17em, height=13em]{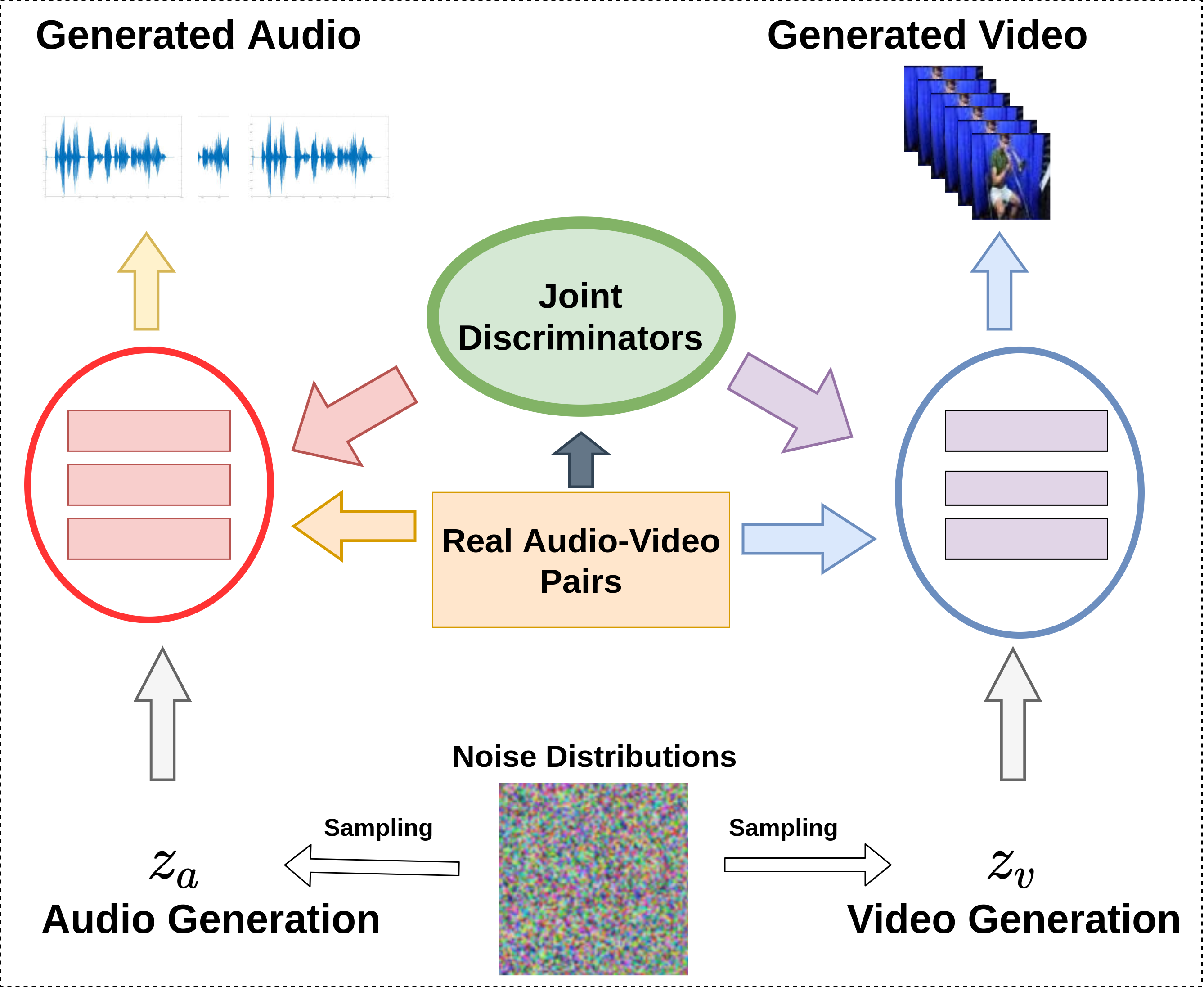}
	\caption{Illustration of collaborative generation of audio-video pairs using joint discriminators.}
	\label{fig:intro}
	\vspace{-2em}
\end{figure}

%Various studies have established the strong correlations in human perception of audio and visual stimuli.
In the real world, different modalities of the same data are highly correlated. Video-audio pairs are one such example. Recent work has extensively explored the generation of data such as audio, video (without the corresponding audio), and text. %Modeling and being able to generate these realistically is always a challenge. %Several works solve for generative modeling of images, audio signals, and videos (without the corresponding audio). 
These methods consider the single modality generation problem either in the form of unconditional generation or conditional generation. Meta information provided for conditional generation can be the class label or one of the other modalities.
% The diConditions are provided in terms of meta information; it could be the class label or other modality. 
Cross model generation focuses on the latter form of conditional generation. The former is known as single modality generation, such as audio generation or video generation, which generate the required modality independently. One of the major drawbacks of these methods is that they do not consider that audio and video streams of the same data are highly correlated and could be generated jointly both benefiting each other. Most multimedia synchronization methods developed in the past have dealt with uni-modal streams
and consider only audio data or video data.  Recently, there have been a few works~\cite{chen2017deep,hao2017cmcgan} that aim to solve cross-modal generation. However, they generate a single image frame for an audio duration and hence fail in real-world video generation as they do not generate a sequence of image frames (i.e. a real video signal) corresponding to the synchronized audio.
The interaction between audio and vision, despite seeing traction as of late, is
still largely unexplored. %This is a particularly relevant topic to the vision community because humans routinely perform tasks which involve multiple modalities. 
In order to solve the problem effectively, we propose a collaborative learning framework that uses adversarial models to ensure that there is out of sample generalization. Specifically, we incorporate multiple discriminators to ensure that the samples generated individually for each modality are realistic. Figure~\ref{fig:intro} shows our proposed framework.
% We further incorporate cross-modal discriminators that ensure correspondence between the individual distributions. 

The exact problem that we want to tackle has not been explored precisely and therefore we did not have a dataset directly applicable to our task. Hence, we composed two novel datasets - Hand Motion Music Dataset (HMMD) and Sub-URMP-video dataset (the latter by modifying an existing dataset).  These datasets are primarily related to musical instruments, and through these datasets, we explore the accurate joint-modal correspondence between audio and video signals. To summarize, we make the following main contributions in this work:
1) We propose a joint collaborative audio-video generation module that uses noise as a latent vector. 2) We provide an audio-video multimodal dataset comprising specifically chosen videos where audio is a direct outcome of the motion (actions) happening in the video frames.
3) We formulate the joint audio-video generation training problem in an adversarial framework and achieve promising results.
\section{Related work}
\vspace{-0.5em}
% \textbf{Video Generation}
Generative adversarial networks~\cite{goodfellow2014generative} is one of the unsupervised techniques that maps low dimensional latent features to high dimensional data. Various architectures have been suggested for generating images using GANs~\cite{karras2017progressive,radford2015unsupervised,huang2018multimodal,wan2019towards}. These image generation architectures have also been extended for video generation. MoCoGAN~\cite{tulyakov2018mocogan} proposes to solve the video generation problem by decomposing motion and content vector. The work in generation is not only limited to the visual modality, but it also extended for other modalities such as music~\cite{huang2018music,engel2019gansynth} and speech generation~\cite{ephrat2018looking,harwath2018jointly,donahue2018exploring}. For instance, SEGAN~\cite{pascual2017segan} uses and adversarial framework to generate speech signals. Van et al.~\cite{vanwavenet} introduced a WaveNet model based on an autoregressive method to predict raw audio samples. Engel et al.~\cite{engel2017neural} used an auto-encoder model in WaveNet to generate musical instrument sounds. Other multi-modal generation frameworks that have been proposed in the space of audio-visual modalities are ~\cite{chen2017deep,hao2017cmcgan,zhou2017visual,lee2018conditional}. Most of the above audio generation works are based on conditional generation conditioned on an image frame of the corresponding video. Recently, several approaches that solve for the alignment of various modalities~\cite{arandjelovic2017objects,senocak2018learning,gao2018learning,ephrat2018looking,owens2018audio,harwath2018jointly,oh2019speech2face,owens2016visually} have also been suggested. Music gesture~\cite{gan2020music}  uses the keypoint-based structured representation to explicitly model the body's and finger's dynamics motion cues for visual sound separation. A few very recent works have also explored the multimodal generation problem. Gan et al.\cite{gan2020foley} synthesized plausible music for a silent video clip of people playing musical instruments. Another similar work~\cite{su2020audeo} generated music for a given video. \cite{chen2020generating} generated visually aligned sounds from videos. In sound2sight~\cite{cherian2020sound2sight}, future video frames and motion dynamics are  generated by conditioning on audio and a few past frames. In SA-CMGAN~\cite{tan2020spectrogram} self-attention mechanism is applied to cross-modal visual-audio generation.

%   Deep neural networks have also been applied to  and audio \cite{owens2018audio,harwath2018jointly} processing.   \cite{chung2014empirical} and~\cite{mehri2016samplernn} use the recurrent autoregressive method to predict raw audio samples. These autoregressive methods produce higher audio fidelity than WaveGAN~\cite{donahue2018adversarial}. The work by~\cite{chen2017deep,hao2017cmcgan,zhou2017visual,lee2018conditional} led to advancements in deep learning-based methods for conditional audio generation problem. Most of the above audio generation works are based on conditional generation based on the image. Recently, several works that solve for the alignment of various modalities~\cite{arandjelovic2017objects,senocak2018learning,gao2018learning,ephrat2018looking,owens2018audio,harwath2018jointly,oh2019speech2face}  have been suggested. In the field of music generation,~\cite{dieleman2014end, zhu2016learning} have proposed frequency selective filter banks based method for generating audio in the human hearing range. Huang et al. ~\cite{huang2018music} have proposed LSTM based model for music generation. In this paper, we propose a GAN based method to generated video and audio jointly leading to naturalistic images.

\begin{figure*}[!ht]
	\centering
	\includegraphics[width=0.790\textwidth,height=0.405\textwidth]{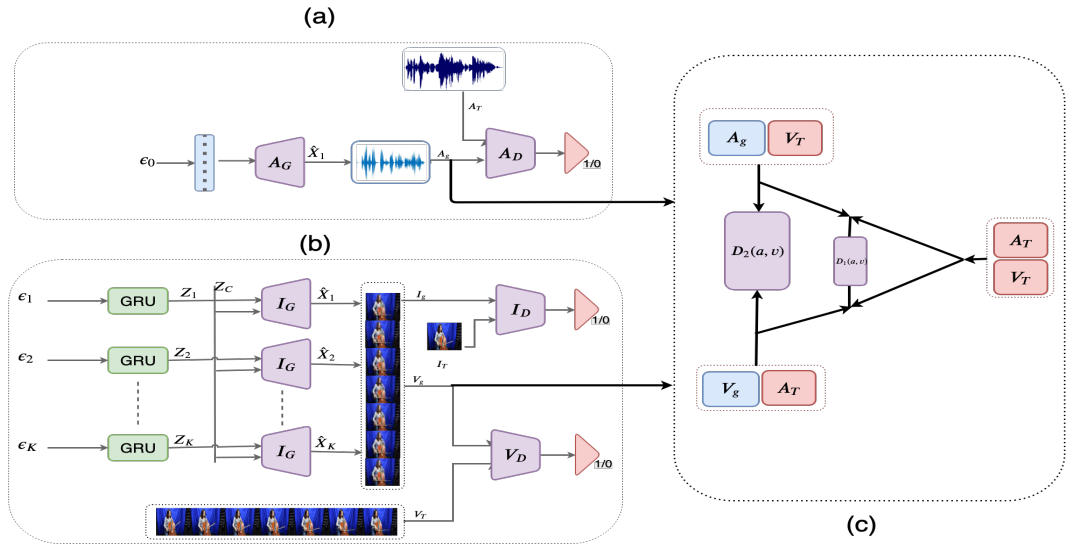}
	\caption{In our proposed model, audio and video are generated using a latent noise vector. The joint discriminators are trained on joint distributions of audio-video pairs.}
	\label{fig:main}
\end{figure*}

\section{Method}
\vspace{-0.71em}

 The proposed framework consists of three modules - audio generation module, video generation module and joint discriminators module.
 
 \noindent\textbf{Audio Generation Module:}
It consists of an audio generator $A_G$, that generates audio from noise $\epsilon_0$ and a discriminator~\cite{donahue2018adversarial}.  This generator module is parameterized through a function $G_{a}(\epsilon_0, W_{ga})$ where $W_{ga}$ are the weights of the model. We work with the raw audio waveform, and the generator produces raw waveforms instead of spectrograms. The generator module uses 1-D deconvolutional layers. Since there are frequency bands present in each audio waveform, unlike the image signal where the frequency patterns are not so common, an image discriminator architecture doesn't work well for audio generation. The discriminator easily learns the frequency pattern commonly known as the checker board effect and discards those waveforms.
% This makes the task of the discriminator more challenging.
These artifact frequencies occur for a particular phase. Therefore, our discriminator needs to learn this trivial policy to discard such generated examples.  To resolve this hurdle, we employ the skill of phase shuffle operation~\cite{donahue2018adversarial} to prevent our audio discriminator from learning such artifacts. It is parameterized through a function $D_{a}(\epsilon_0, W_{da})$ where $W_{da}$ are the weights of the model. Generator $G_a$ is trained to minimize the following value function, while $D_a$ is trained to maximize it.
\vspace{-1em}
 \begin{equation}
 \begin{split}
%  \resizebox{.9 \textwidth}{!} 
     V(D_{a}, G_{a})=\mathbb{E}_{\boldsymbol{a} \sim R_{\boldsymbol{A}}}[\log D_{a}(\boldsymbol{a})]+\\
     \mathbb{E}_{\boldsymbol{\epsilon} \sim P_{\epsilon}}[\log (1-D_{a}(G_{a}(\boldsymbol{\epsilon_0})))]
 \end{split}
  \vspace{-0.75em}
 \end{equation}

   $\boldsymbol{a}$ is sampled from real data($R_{\boldsymbol{A}}$) and $\boldsymbol{\epsilon} $ is random noise.

\begin{table*}[!ht]
\centering
\caption {Sub-URMP Video dataset} 
% \vspace{-1em}
\scalebox{0.8}{
\begin{tabular}{cccccccccccccc}
\hline
{Cat.}& Cello & D.Bass & Oboe & Sax & Trumpet & Viola &  Bassoon & Clarinet & Horn & Flute &Trombone& Tuba& Violin  \\ 
\hline
Train   & 881  & 123  &394  & 696  & 100  &648&135& 716  & 475  &511  & 787  & 323  &707  \\ 
Test  & 94  & 114 & 36& 81  & 49  &48  &37& 92  & 49 & 90& 77  & 49  &91  \\ 
% \hline 
% Stack Attention & 83.40 & 79.72  &  58.82 & 56.12  & 69.52\\
\hline
\end{tabular}
}
\label{tab:dataset_URMP}
\vspace{-1em}
 \end{table*}
 
 \begin{table}[ht]
\vspace{-1em}
\centering
\caption {Hand Motion Music Dataset(HMMD). Complete videos refer to raw video clips collected from Youtube for composing our HMMD dataset detailed in Section~\ref{sec:dataset}. We had an 80:20 train-test split for the composed dataset. AG:Acoustic Guitar} 
% \vspace{-1em}
\scalebox{0.8}{
\begin{tabular}{cccccccc}
% \begin{tabular}{p{1cm}p{0.8cm}p{0.8cm}p{0.8cm}p{0.8cm}p{0.8cm}p{0.8cm}p{0.8cm}}

\hline
{Category }& Cello & Tabla & Ghatam & Erhu & AG & Xylophone & Violin  \\ 
\hline
 Videos  &15   & 3  & 1& 12 &  11 &10  &25  \\ 
  Frame & 1251  & 317 & 454& 519 & 1041  & 800 & 1847 \\ 
% \hline 
\hline
\end{tabular}
}
\label{tbl:dataset1}
\vspace{-1em}
 \end{table}

 \noindent\textbf{Video Generation Module:}
  Our video generation component takes inspiration from the idea of decomposing videos into a separate motion and content part~\cite{tulyakov2018mocogan}. Our video clips $v_n= [I_1,\dots,I_n]$ are composed of $N$ frames.
%  We generate our video clips using spatial-temporal CNN based video generator $G_V$ and video discriminator $D_v$. 
%  The generator $G_I$ maps random sequential noise vectors $\epsilon_N= [\epsilon_1,\dots,\epsilon_N]$ to fixed-length sequential video frames $v_n= [I_1,\dots,I_N]$. The discriminator $D_v$ is trained to distinguish the real video clip from the generated video clips. 
%  \textit{i.e} $D_v(v_n)$ =1, if $v_n$ is sampled from the real distribution and $D_v(\hat{v}_N)$ =0, if $v_n$ is sampled from the generated video distribution. . 
The temporal relationship between the video frames is obtained by modeling the latent space vector $Z_n$. The content of the video clips is encapsulated by introducing a context vector $Z_C$ and is combined with every motion vector.  Each input to the video generator is a combination of the latent motion vector and the latent content vector. 
% We model a sequential network Gated Recurrent Units (GRU) to capture temporal relation present in the video with noise as an input to the network. The output of the GRU network combine with $Z_c$ to get final latent input $Z_I$
% \begin{equation}[Z_{I}(1),\dots,Z_{I}(N)]=\Big[\begin{bmatrix}
% Z_c  \\
% Z_1 
% \end{bmatrix},\dots,\begin{bmatrix}
% Z_c  \\
% Z_N 
% \end{bmatrix} \Big]
% \end{equation}
Each $Z_I$ is an input to the image generator to generate each frame in the video clip. We use individual image discriminator $D_I$ and a video discriminator $D_V$ to play the role of a judge to provide criticism to video (image) generator $G_I$.
% The role of the image discriminator $D_I$ is to criticize the image generator $G_I$ on every image frame. 
% The discriminator is trained to identify if the frame is sampled from a generated video clip or the real video clips. 
% Moreover, the video discriminator takes a fixed number of frames and determines if the frames are sampled from real video clip or generated one $\hat{V}_N$.
% The generated video is given by :
% \begin{equation}
% \hat{v}=\Big [ G_I\Big(\begin{bmatrix}
% Z_c  \\
% Z_1 
% \end{bmatrix}\Big)
% % \quad
% ,G_I\Big(
% \begin{bmatrix}
% Z_c  \\
% Z_2 
% \end{bmatrix}\Big)
% \dots
% G_I\Big(\begin{bmatrix}
% Z_c  \\
% Z_N 
% \end{bmatrix}\Big)\Big]
%  \end{equation}
 The learning problem is formulated as:
%  \vspace{-1.5em}

% \begin{equation}
% \begin{split}
% \min_{G_I} \max_{D_I, D_V} ( C_v(D_I, D_V, G_I))
% \end{split}
% \vspace{-2.5em}
% \end{equation}
 \vspace{-1.8em}
% The cost function  $C_v(D_I, D_V, G_I)$ is defined
\begin{equation*}
\begin{split}
& \min_{G_I} \max_{D_I, D_V} ( C_v(D_I, D_V, G_I))=  \\ 
& \mathbb{E}_{v \sim  \mathcal{V}} [\log D_{I}(S_{1}(v))] + \mathbb{E}_{\hat{v} \sim G_V} [\log (1- D_{I}(S_{1}(\hat{v})))] + \\
&\mathbb{E}_{v \sim  \mathcal{V}} [\log D_{V}(S_{T}(v))] + \mathbb{E}_{\hat{v} \sim G_V} [\log (1- D_{V}(S_{T}(\hat{v})))]
\vspace{-0.5em}
\end{split}
\end{equation*}

 $S_1$ and $S_T$ are two sample functions for image discriminator $D_I$ and video discriminator $D_V$  which takes as input, one frame and T consecutive frames respectively. The sample functions $S_1$, and $S_T$ are sampled from real $ \mathcal{V}$ and generated video $\hat{V}$ clip. $\hat{v}$ is the generated video.

\noindent \textbf{Joint Discriminators:}\label{joint model}
 The marginal distributions are represented  for audio and video as $p(a)$ and $p(v)$. The joint distribution of audio and video is denoted by $p(a,v)$. To jointly learn the audio-video generation, we use two joint discriminators, inspired by~\cite{gan2017triangle}. This pair of joint discriminators is trained to distinguish real data pairs $(a,v)$ and two kinds of generated data pairs  $(a,\hat{v})$, $(\hat{a},v)$. The main objective of this module is to match the three joint distributions: $p(a,v), p(a, \hat{v}) \& p(\hat{a}, v)$ such that the generated fake data pairs $(a,\hat{v})$ and $(\hat{a}, v)$ are indistinguishable from the real data pairs $(a, v)$. We follow an adversarial framework to match the joint distributions.  The overall-cost function for joint discriminators is defined as:
%  \vspace{em}
\begin{equation}
\vspace{-1em}
\begin{split}
\small
&\min_{G_A,G_I} \max_{D_1, D_2}  C_M(D_1, D_2, G_A,G_I) 
=\mathbb{E}_{a \sim \mathcal{A},v \sim  \mathcal{V} } [\log D_{1}(a,v)]\\
& + \mathbb{E}_{\hat{a} \sim G_{A},v \sim  \mathcal{V}} \Big[\log \Big( (1- D_{1}(\hat{a},v)  D_{2}(\hat{a},v) \Big) \Big] \\
& + \mathbb{E}_{{a} \sim  \mathcal{A},\hat{v} \sim G_I} \Big[\log \Big( (1- D_{1}(a,\hat{v}) (1- D_{2}(a,\hat{v}) \Big) \Big]
\end{split}
\vspace{-1em}
\end{equation}

% \begin{equation}
% \begin{split}
% \small
% &\min_{G_A,G_I} \max_{D_1, D_2}  C_M(D_1, D_2, G_A,G_I)= \\
% & \mathbb{E}_{a \sim R_{A},v \sim R_{V} } [\log D_{1}(a,v)]
% + \mathbb{E}_{\hat{a} \sim G_{A},v \sim R_V} \Big[\log \Big( (1- D_{1}(\hat{a},v) \dot D_{2}(\hat{a},v) \Big) \Big] \\
% & + \mathbb{E}_{{a} \sim R_{A},\hat{v} \sim G_I} \Big[\log \Big( (1- D_{1}(a,\hat{v}) \dot (1- D_{2}(a,\hat{v}) \Big) \Big]
% \end{split}
% \end{equation}

Here, $ \mathcal{A}$ and $ \mathcal{V}$ are the audio and video data distributions respectively. The objective of $D_1$  is to  distinguish whether a sample pair is from real $p(a, v)$ or not, If the sample pair is not from $p(a, v)$, we use another discriminator $D_2$ to distinguish if the sample pair is from $p(\hat{a}, v)$ or $p(a,\hat{v})$. Both the discriminators $D_1 \& D_2$ work collectively, and the use of both of them implicitly define a ternary discriminative function $D$. 

%comment by badri
% \begin{figure}[ht]
% 	%\vspace{1in}
% 	\centering
% 	\includegraphics[width=0.45\textwidth]{fig/hmmd_dataset2.png}
% 	\caption{Hand Motion Music Dataset: Sample examples of all the instrument present in our dataset.}
% 	\label{fig:our_data_org}
% \end{figure}

%  \begin{figure}%[!h]
%  \small
%  \centering
%  \begin{tabular}[b]{ c}
% %  (a) Baseline: Audio generated from Noise \\
% \includegraphics[width=0.38\textwidth]{fig/audio_acm_hmmd_independent.png}\\
% %  (b) Joint : Audio generated from Noise with joint discriminator\\ 
% % \vspace{-1em}
% \includegraphics[width=0.38\textwidth]{fig/audio_acm_hmmd_joint.png}
%   \end{tabular}
% \caption{Spectrograms - Upper Graph: This graph shows how audio frequency varies over time for generations from the baseline model in case of HMMD dataset. Lower Graph: This graph illustrates the variation of audio frequency over time for the joint model for HMMD dataset. The generated audio files have been shared in the supplementary material.}
% % 	  \vspace{-0.7cm}
%   \label{fig:HMMD_spet_gen}
% %   \vspace{-2em}
% \end{figure}

 \vspace{-0.5em}
\section{Experiments}
\vspace{-1.5em}
\subsection{Datasets}\label{sec:dataset}
\vspace{-0.5em}
We compose two novel datasets to train and evaluate our models- Sub-URMP-Video Dataset, a video version of the Sub-URMP dataset~\cite{chen2017deep} and Hand Motion Music Dataset (HMMD). To show the effectiveness of our method, we need videos with significant movement across frames and where audio is directly related of the motion happening in the video frames. We construct our own dataset (HMMD) comprising videos from Youtube focusing particularly on percussion instruments like Tabla, etc. which have significant hand movements across frames. Both the datasets and the code will be made publicly available.
% for reproducibility of the paper.
% \vspace{-1em}

\noindent\textbf{Sub-URMP-Video Dataset:}
It contains 13 musical instrument categories. It contains image-audio pairs with ten images per second and audio of 0.5 seconds paired with each image. We form 10 frames per second(fps) video with 1 second of audio and match them to make it suitable for our task. We also manually remove the videos that contain different speakers.
% Additionally, we remove the pairs that have audio's loudness below a threshold(-35dBFS). 
The basic information is summarized in Table~\ref{tab:dataset_URMP}.

\noindent\textbf{HMMD:}
There are seven types of instruments, in total and 77 video clips. The number of clips ranges from 1 to 25 for various instruments. We preprocess the dataset to form videos of 1 sec each having a frame rate of 25 per sec. Audio streams are sampled at 16kHz. The proposed Hand Motion Music Dataset consists of 6229 video-audio pairs. The details of the dataset are present in Table~\ref{tbl:dataset1}. 
% Some sample examples of the HMMD dataset are presented in supplementary at link~\footnote{https://delta-lab-iitk.github.io/AVG/}.
% \footnote{https://drive.google.com/file/d/1fyAI1HPzzB18WZM7EwL_yy6BnDJUAW36/view?usp=sharing}
% \footnote{https://drive.google.com/file/d/1fyAI1HPzzB18WZM7EwL_yy6BnDJUAW36/view?usp=sharing}. 
% We evaluated the proposed model on two datasets: Sub-URMP-Video Dataset and Hand Motion Music Dataset (HMMD) as shown in Table~\ref{tab:dataset_URMP} and  Table~\ref{tbl:dataset1}. We also compare the model with baseline model in terms of inception score,  Statistically-Different Bins, pitch entropy, Frechet Inception Distance (FID) for audio and qualitative performance, along with a human evaluation for video and audio. Please note that baseline model refers to our own model without the joint discriminator.
% We have split the video and audio samples into one second's duration for both the dataset. The detailed algorithm is shown in the Algorithm.~\ref{algo}

\vspace{-1em}
\subsection{Results}
\vspace{-0.5em}

% We evaluated the proposed model on two datasets: Sub-URMP-Video Dataset and Hand Motion Music Dataset (HMMD).
% as described in Table~\ref{tab:dataset_URMP} and  Table~\ref{tbl:dataset1}.
We compare our model with the baseline model using both quantitative and qualitative methods such as inception score,  Statistically-Different Bins, pitch entropy, Frechet Inception Distance (FID) for audio and qualitative performance, along with a human evaluation for video and audio.
We work with two modalities having separate architecture for generation. The video generation module is inspired from MoCoGAN~\cite{tulyakov2018mocogan}, and for audio generation model, it is adapted from WaveGAN~\cite{donahue2018adversarial}. For video generation, we used a  deconvolution layers based generator for producing 64x64 images from a 100-dim noise vector. To the best of our knowledge, there was no prior work in joint audio video generation as has been iterated in the paper and therefore the absence of a precise baseline. Consequently, we remove the joint discriminator component (c) from our model and consider the remaining model as a baseline for quantitative evaluations. and compare the effect of joint generation to isolated generations.
Therefore, the baseline model we choose are the single modality generation components for audio and video respectively.
The other details are provided in project page~\footnote{https://delta-lab-iitk.github.io/AVG/}.

 \begin{table}[ht]
  \caption{Audio Evaluation  Metrics for proposed models. \newline BL: Baseline, JM:Proposed Joint Model}
%   \vspace{-1em}
%   \setlength\tabcolsep{1.5pt}%
\scalebox{0.8}{
\centering
%   \begin{tabular}{llllll}
  \begin{tabular}{p{2.2cm}p{1cm}p{1cm}p{1cm}p{0.7cm}p{0.5cm}}
    % \toprule
        \hline
    Dataset-Model &Human($\uparrow$) & NDB($\downarrow$) & FID($\downarrow$) & IS($\uparrow$) & PE($\downarrow$)  \\
    % \midrule
    \hline
    S-URMP(BL) & 3.2 & 14&111.25&1.38 &7.53\\
    S-URMP(JM) &3.6 & 10& 97.43&  1.44 &6.12\\ \hline
    HMMD(BL) &3.1 &36 &93.00&1.34&7.33\\
    HMMD(JM &3.4 &29&79.89&1.48& 5.42 \\\hline
    %  \bottomrule
\end{tabular}
}
\label{tbl:audio}
\vspace{-1em}

\end{table}
\vspace{-1em}
\begin{table}[!hbt]
  \caption{Video Evaluation  Metrics for the proposed models}
\centering
  \label{tab:freq_video}
  \scalebox{0.8}{

  \begin{tabular}{lcccc}
    % \toprule
    \hline
    Dataset&Model &Human($\uparrow$)  &  IS($\uparrow$)   \\
    % \midrule
    \hline
    Sub-URMP&Ground Truth &4.6 & 1.42\\
    Sub-URMP&Baseline   &3.1 &1.88 \\
    Sub-URMP&Joint Model  &3.7 &   1.91 \\ \hline
    HMMD&Ground Truth &4.8 & 1.26\\
    HMMD&Baseline  &2.9 &4.24\\
    HMMD&Joint Model &3.5 &4.33 \\
    %  \bottomrule
    \hline
\end{tabular}
}
 \vspace{-1em}
 \end{table}
 
\subsection{ Audio Performance Evaluation method }
\vspace{-0.5em}
To validate the proposed model, we provide a quantitative analysis of the generated audio signals. The audio signals are generated using only noise as a latent vector, unlike the previously proposed methods~\cite{chen2017deep,hao2017cmcgan} that use a conditional vector for audio generation. We also compare the quality of the audio signal produced without the joint discriminator model. From this assessment, we can clearly say that the joint audio-video model helps to generate better audio samples. 
% The generated audio samples are shown in Figure-~\ref{fig:HMMD_spet_gen}.
\vspace{-1em}

\subsection{Video Performance Evaluation}
\begin{figure}[htb]
	%\vspace{1in}
	\centering
	\includegraphics[width=0.480\textwidth]{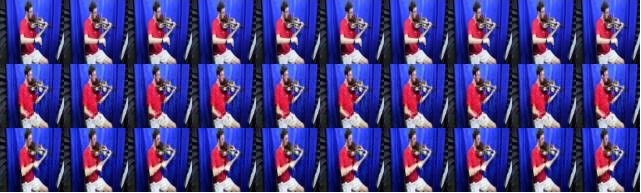}
% 	\vspace{-2em}
	\caption{Results on Sub-URMP Dataset:In this figure, the first row indicates the given ground truth video. The second row indicates generated videos for our baseline model and third row shows generated video through our joint model}
	\label{fig:visu}
% 	\vspace{-1em}
\end{figure}

\vspace{-0.5em}
% \subsubsection{Video Generation}
We also validate our model using the video generation task. The state-of-art video generation model~\cite{tulyakov2018mocogan} is considered as a baseline for our task which is a component of our joint model. We cannot compare our model with other video generation methods because they rely on conditional video generation which is a much easier task. The generated frames are shown in Figure-~\ref{fig:visu} for Sub-URMP dataset. We observe that the joint audio-video generation framework helps the model to learn better video frames. We observe that the inception score improves 0.3\% for Sub-URMP and 0.9\% for HMMD dataset.

\vspace{-1em}
\section{Analysis}
% \vspace{-0.5em}
The joint model outperforms the baseline models in all our evaluations-in Table 2 as well as Table 4. We have designated alongside every evaluation metric with arrows, where up-arrow ($\uparrow$) indicates that a larger value is better and similarly the down-arrow ($\downarrow$) suggests that a lower value is better.
% Here we compare our model with other state of the art models.

\noindent\textbf{{Comparison with Deep Cross model generation:}}
% \subsection{Comparison with Deep Cross model generation}
\noindent In Deep cross-modal generation~\cite{chen2017deep}, audio and image generations are conditional in nature. But in our proposed model, audio and video are generated from a latent noise vector. The former also uses the music class label for generation while our work is completely unsupervised. Another difference is that we are generating videos (a sequence of image frames ), which is a more challenging task, unlike the previous work which generates only images.  %The audio signal always corresponds to the video it is a component of, not the image data.

% In the Deep cross model generation~\cite{chen2017deep} framework, the audio and image generation are conditional. But in the proposed model, we are generating audio and video from the latent noise vector. This work also uses the music class label for the cross model generation while our work is completely unsupervised. Another major difference is that we are generating the videos instead of images which is more challenging task. The audio signal always corresponds the video not the image data. 
\noindent\textbf{{Comparison with CMC GAN}:}
% \subsection{Comparison with CMC GAN}
% The subsequent version of the Deep Cross model generation is CMC-GAN~\cite{hao2017cmcgan}, which uses the cycle consistency between generated audio and video samples. This work also considered as conditional generation of audio and images data. This works also use the image data rather than video. Our proposed work generate the audio and video data without conditioning on each other. 
% \noindent The subsequent version of the Deep Cross model generation is CMC-GAN~\cite{hao2017cmcgan}, which uses the cycle consistency between generated audio and video samples. However, even this work comprises the conditional generation of audio and images data. It faces the same drawback as the previous one like they work on image data rather than video. Our proposed work generates audio and video data without conditioning on each other.
 The CMC-GAN~\cite{hao2017cmcgan} uses the cycle consistency between generated audio and video samples. However, this work too relies on conditional generation of audio and images data. It faces the same drawback as the previous one as it works only on images and not on videos whereas our model generates audio and video modalities collaboratively without conditioning on each other.

\noindent\textbf{{Comparison with Triangle GAN:}}
% \subsection{Comparison with Triangle GAN}
\noindent The Triangle GAN~\cite{gan2017triangle} model too does conditional generation by joint training of the modalities. Again, our work does not use any conditional dependence rather it jointly learns the audio-video generation. 

\noindent\textbf{{Comparison with independent modality Generation model:}}
% \subsubsection{Comparison with independent modality Generation model}
\noindent The other video generation models such as MoCoGAN~\cite{tulyakov2018mocogan} do not consider audio modality for generating videos. Similarly audio generation models such as WaveGAN~\cite{donahue2018synthesizing} also generate audio without considering the video information. Therefore, to the best of our knowledge, there are no other models available that can generate audio and video simultaneously by jointly training the generators.

\noindent \textbf{Comparison with SA-CMGAN:} In SA-CMGAN~\cite{tan2020spectrogram}  audios and images are generated using the conditional GAN, while we generate videos and audio, which is more challenging using the joint learning. They also use the class label while training the model. In contrast to it, proposed framework is completely unsupervised.

% \vspace{-1.5em}

% \vspace{-1em}
% \vspace{-0.5em}
%%%%%%%%%%%%%%%%%%%%%%%%%%%%%%%%%%%%%%%%%%%%%%%%%
\section{Conclusion}
% \vspace{-0.5em}
% This paper proposes a JAVS model for cross-modal visual-audio mutual generation from noise. We have incorporated video genreration and audio generation into a single common network. Further, our model can be trained for end to end generation for better convenience. Numerous experiments have been conducted and our mnodel JAVS achieves significantly good results. Our generated videos and audio are realistic as is confirmed by human evaluation. Moreover, we develop a dynamic multimodal classification network, which can solve the missing modality problem in a true sense and in an effective way. 
This paper proposes a collaborative generation paradigm for audio and video generation without any conditioning involved. It can also be trained for an end to end fashion for better convenience. The generated videos and audio are realistic, as is manifested by numerous experiments as well as human evaluation. 
% Moreover, we develop a dynamic multimodal classification network, which can solve the missing modality problem in a true sense and in an effective way.
The results clearly depict that collaborative generation successfully takes advantage of the subtle relationship that video stream has with the audio stream to generate  naturalistic images and improved quality audio. The proposed two datasets  will immensely help the community in audio-visual research. 
Using collaborative generation of related modalities to improve the quality of individual modalities generated is a promising direction which we have initiated through this work.
Hence, we believe that this paper will act as a catalyst for work in multimodal generation. 

% \vfill\pagebreak

% \section{REFERENCES}
% \label{sec:refs}

% List and number all bibliographical references at the end of the
% paper. The references can be numbered in alphabetic order or in
% order of appearance in the document. When referring to them in
% the text, type the corresponding reference number in square
% brackets as shown at the end of this sentence \cite{C2}. An
% additional final page (the fifth page, in most cases) is
% allowed, but must contain only references to the prior
% literature.

{
\small
\bibliographystyle{IEEEbib}
\bibliography{strings,refs}
}
\end{document}